\documentclass[letterpaper, 10 pt, conference]{ieeeconf}  % Comment this line out if you need a4paper

\IEEEoverridecommandlockouts                              % This command is only needed if 
\usepackage{graphics} % for pdf, bitmapped graphics files
\usepackage{amsmath} % assumes amsmath package installed
\usepackage{cite}
\usepackage{subfigure}
\usepackage{tabulary}
\usepackage{multirow}
\usepackage{todonotes}
\graphicspath{{figs/}}
\usepackage[colorlinks,linkcolor=blue,anchorcolor=blue,citecolor=red]{hyperref} 
\usepackage{wrapfig,lipsum,booktabs}
\usepackage{xcolor}
\usepackage{color}
\usepackage{cite}
\usepackage{algorithmic} %format of the algorithm 
\usepackage[ruled,vlined,commentsnumbered]{algorithm2e}

\usepackage[flushleft]{threeparttable}
\usepackage{gensymb}  % degree symbol
\usepackage{amsmath}
\usepackage{amssymb}

\usepackage{flushend}

\graphicspath{{figs/}}

\usepackage{siunitx}
\usepackage{url}   % url link

\DeclareMathOperator*{\argmax}{arg\,max}

\title{\LARGE \bf
Pop-up SLAM: Semantic Monocular Plane SLAM\\for Low-texture Environments
}

\author{Shichao Yang, Yu Song, Michael Kaess, and Sebastian Scherer % <-this % stops a space
\thanks{The Robotics Institute, Carnegie Mellon University, 5000 Forbes Ave, Pittsburgh, PA 15213, USA.
		{\tt\small \{shichaoy, songyu, kaess, basti\}@andrew.cmu.edu}}
}

\begin{document}

\maketitle
\thispagestyle{empty}
\pagestyle{empty}

%%%%%%%%%%%%%%%%%%%%%%%%%%%%%%%%%%%%%%%%%%%%%%%%%%%%%%%%%%%%%%%%%%%%%%%%%%%%%%%%
\begin{abstract}
    % importance of research
% Problem descripttion, scope of research
% Approach
% Results
Existing simultaneous localization and mapping (SLAM) algorithms are not robust in challenging
low-texture environments because there are only few salient features. 
The resulting sparse or semi-dense map also conveys little information for motion planning. 
Though some work utilize plane or scene layout for dense map regularization, they require 
decent state estimation from other sources.
In this paper, we propose real-time monocular plane SLAM to demonstrate that scene understanding 
could improve both state estimation and dense mapping especially in low-texture environments. The 
plane measurements come from a pop-up 3D plane model applied to each single image. We also combine 
planes with point based SLAM to improve robustness.
On a public TUM dataset, our algorithm generates a dense semantic 3D model with pixel 
depth error of 6.2 cm while existing SLAM algorithms fail. On a 60 m long dataset 
with loops, our method creates a much better 3D model with state estimation error of 0.67\%.

%To improve the To  the ill-constrained problem of plane SLAM, we combine it with 
%a state-of-art point based SLAM method in two variants to improve the robustness.
%A 3D plane world is firstly generated from
%We first utilize a Convolutional Neural Network to create a pop-up 3D plane world. These planes are then 
%treated as plane measurements in SLAM to provide both state estimates and dense maps.
%The experiments show the validity of our approach. 

%We propose a real time semantic plane SLAM from monocular images 
%by incorporating scene layout understanding, especially suitable for low texture
%environments. Scene understanding has been researched for decades, while few of 
%papers extend it to multiple images in SLAM framework. Some existing SLAM methods 
%only utilize plane information in the post-processing to create dense maps, assuming state 
%estimation is given from other sources. Furthermore, they cannot provide a semantic label 
%of planes.

\end{abstract}

\section{Introduction}
% what is the topic why this topic is important
% brief intro of existing methods
% how our method
Simultaneous localization and mapping (SLAM) is widely used
for tasks including autonomous navigation, 3D mapping and inspection.
Various sensors can be used for SLAM such as laser-range finders 
cameras, and RGB-D depth cameras. Monocular cameras are a popular choice of sensor on robots as they can 
provide rich visual information at a small size and low cost. They are especially
suitable for weight constrained micro aerial vehicles that can carry only one camera.
Therefore, in this work we focus on using monocular images to estimate the pose and map of the environment.

On one hand, many existing visual SLAM methods utilize point features such as
direct LSD SLAM \cite{engel2014lsd} and feature based ORB SLAM \cite{mur2015orb}.
These methods track features or high-gradient pixels across frames to find correspondences
and triangulate depth. They usually perform well in environments with
rich features but cannot work well in low-texture scenes as often found in corridors.
In addition, the map is usually sparse or semi-dense, which does not convey much information
for motion planning.

On the other hand, humans can understand the layout, estimate depth and detect obstacles
from a single image. Many methods have been proposed to exploit the geometry cues and 
scene assumption in order to build a simplified 3D model. Especially in recent years, 
with the advent of Convolutional Neural Networks (CNN) \cite{krizhevsky2012imagenet}, 
performance of visual understanding has been greatly increased.

\begin{figure}[t]
  \centering
   \includegraphics[scale=0.42]{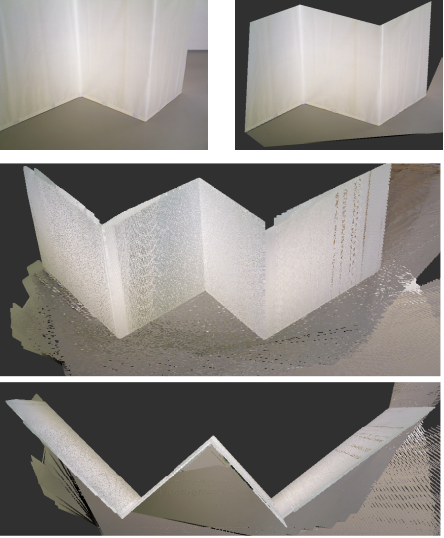}
   \caption{3D reconstruction on low-texture TUM dataset. (top) Single frame, raw image on left and 3D pop-up plane model on right.
   (center) Dense reconstruction using our \textit{Pop-up Plane SLAM}. Each plane 
   has a label of either a specific wall or ground. (bottom) Top view of the 3D model. Existing state-of-art
   SLAM algorithms fail.}
   \label{fig:method overview}
\end{figure}

In this paper, we combine scene understanding with traditional v-SLAM to increase the 
performance of both state estimation and dense mapping especially in low-texture environments.
We use a single image pop-up plane model \cite{chao2016pop} to generate plane landmark measurements in SLAM. With 
proper plane association and loop closing, we are able to jointly optimize scene layout
and poses of multiple frames in the SLAM framework. In the low-texture environment of Figure
\ref{fig:method overview}, our algorithm can still generate dense 3D models and decent state estimates
while other state-of-the-art algorithms fail. However, plane SLAM can easily be under-constrained, hence we propose to combine it with traditional point-based LSD SLAM \cite{engel2014lsd} to increase robustness.

In summary, our main contributions are:
\begin{itemize} 
    \item  A real-time monocular plane SLAM system incorporating scene layout understanding,
    \item  Integrate planes with point-based SLAM for robustness,
	\item  Outperform existing methods especially in some low-texture environments and
           demonstrate the practicability on several large datasets with loops.
\end{itemize}

In the following section, we discuss related work. Section \ref{sec:single image pop}
describes the single image layout understanding, which provides plane measurements for plane
SLAM. In Section \ref{sec:plane slam}, we introduce the \textit{Pop-up Plane SLAM} formulation
and combine it with LSD SLAM in Section \ref{sec:point plane SLAM}. Experiments
on a public TUM dataset and actual indoor environments are presented in Section 
\ref{sec:experiments}. Finally, we conclude in Section \ref{sec:conclusions}.

%%%%%%%%%%%%%%%%%%%%%%%%%%%%%%%%%%%%%%%%%%%%%%%%%%%%%%%%%%%%%%%%%%%%%%%%%%%%%%%%

\section{Related Work}
\label{sec:related}
Our approach combines aspects of two research areas: single image scene understanding
and multiple images visual SLAM. We provide a brief overview of these two area.

\subsection{Single Image}
There are many methods that attempt to model the world from a single image. Two representative examples
are cuboidal room box model proposed based on vanishing point by Hedau et al.~\cite{hedau2009recovering}
and fixed building model collections based on line segments by Lee et al.~\cite{lee2009geometric}.
Our previous work \cite{chao2016pop} proposed the pop-up 3D plane model, combining CNNs with geometry
modeling. Results show that our work is more robust to various corridor configurations and lighting conditions
than existing methods.
%the Make3D system \cite{saxena2009make3d}, data driven primitives \cite{fouhey2013data},
%room layout \cite{hedau2009recovering}, building model collections  

\subsection{Multiple Images}
\subsubsection{v-SLAM using points}
Structure from Motion and v-SLAM have been widely used to obtain 3D reconstructions 
from images \cite{klein2007parallel}. These methods track image features across 
multiple frames and build a globally consistent 3D map using optimization.
Two representatives of them are direct LSD SLAM \cite{engel2014lsd} and feature-
based ORB SLAM \cite{mur2015orb}. But these methods work poorly in low-texture environments 
because of the sparse visual and geometric features.

\subsubsection{v-SLAM using planes}
Planes or superpixels have been used in \cite{furukawa2010accurate, concha2015dpptam, pinies2015dense} to provide dense mapping in low-texture areas. But they assume 
camera poses are provided from other sources such as point based SLAM, which may not
work well in textureless environments as mentioned above. Recently, Concha et al.~\cite{concha2015incorporating} also propose to use room layout information to 
generate depth priors for dense mapping however they don't track and update the room layout 
thus can only work in small workspace.

\subsubsection{Scene understanding}
Some works focus on the scene understanding using multiple images, especially in a 
Manhattan world. Flint et al.~\cite{flint2011manhattan} formulate it as Bayesian 
framework using monocular and 3D features. \cite{tsai2011real, furlan2013free}
generate many candidate 3D model hypotheses and subsequently update their probability 
by feature tracking and point cloud matching. Unfortunately, these methods do not use a
plane world to constrain the state estimation and thus cannot solve the problem 
of v-SLAM in low-texture environments.

\section{Single Image Plane Pop-up}
\label{sec:single image pop}
This section extends our previous work \cite{chao2016pop} to create a pop-up 3D plane model 
from a single image. We first briefly recap the previous work, discuss its limitation, and
propose two improvements accordingly.

\subsection{Pop-up 3D Model}
\label{sec:Pop 3D Model}
\begin{figure}[thpb]
  \centering
   \includegraphics[scale=0.38]{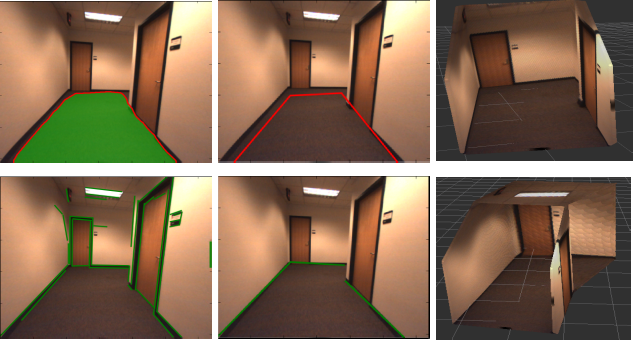}
   \caption{Single image pop-up plane model. (top) Original method of \cite{chao2016pop}. 
   From left to right: CNN segmentation, polyline fitting, pop-up 3D model. (bottom) 
   Improved method. From left to right: line segment detection, selected ground-wall 
   edges, pop-up 3D model. Better ground edge detection and camera pose estimation lead to a more accurate 3D model.}
   \label{fig:pop_improve}
\end{figure}

There are three main steps in \cite{chao2016pop} to generate 3D world: CNN ground
segmentation (optionally with Conditional Random Field refinement), polyline fitting, 
and pop up 3D plane model. It outperforms existing methods in various dataset 
evaluation. However, there are some limitations:

Firstly, \cite{chao2016pop} fit polylines along the detected ground region which 
might not be the true wall-ground edges and thus generate a invalid 3D scene model.
For example in Figure \ref{fig:pop_improve}, it cannot model the right turning hallway.
This results in problems attempting to use these planes in SLAM framework 
because even in adjacent frames, the fitted line segments may be different. However, SLAM requires the 
landmark (in our case, planes), to be invariant across frames.

Secondly, \cite{chao2016pop} uses a zero rotation pose assumption, which in most
cases, is not satisfied. Different rotation angles may generate different pop-up 3D model.

In the following two sections, we solve these problems and generate a more accurate 3D map 
shown at the bottom of Figure \ref{fig:pop_improve}.

%The improved method utilizes true boundaries and camera pose, thus 

\subsection{Optimal Boundary Detection}
\label{sec:Optimal boundary detection}
Instead of using a fitted polygon as a wall-ground boundary, we propose to detect the 
true ground-wall edges. We first extract all the line segments using \cite{von2008lsd}.
But as other line detectors, this algorithm also has detection noise. For example, a long straight line 
may be detected as two disconnected segments. We propose an algorithm to optimally select 
and merge edges as a wall-ground boundary shown in the bottom center of Figure \ref{fig:pop_improve}.

Mathematically, given a set of detected edges $V=\lbrace e_1,e_2,...,e_n \rbrace$,
we want to find the optimal subset edges $S \subseteq V$, such that:
\begin{equation}
    \label{unary_def}
    \max_{S \subseteq V} F(S), \  st\colon S \in I
\end{equation}
where $F$ is the score function and $I$ is the constraint. Due to the complicated scene structures 
in the real world, there is no standard way of expressing $F$ and $I$ as far as we know, so we intuitively design them
to make it more adaptable to various environments, not limited to a Manhattan world as it is typically done in many current approaches \cite{concha2015incorporating} \cite{tsai2011real}.
% based on the rough ground segmentation by CNN \cite{chao2016pop}.
%Since we already have a rough ground segmentation by CNN \cite{chao2016pop}, 
%we can design $F$ and $I$ based on that.
%\cite{tsai2011real} randomly chooses three edges in difference orientations to generate 3D model while 
%\cite{hedau2009recovering} generates ground-wall boundary from vanishing points.

The first constraint indicates that edges should be close to the CNN detected boundary curve $\xi$ 
within a threshold shown as red curve in the top left of Figure 
\ref{fig:pop_improve}. It can be denoted as:
\begin{equation}
\label{eq:edge constraint 1}
I_{close}=\lbrace S\colon \forall e \in S, \ dist(e, \xi)< \delta_{close} \rbrace
\end{equation}

\begin{figure}[h]
  \centering     %%% not \center
  \subfigure[]
    {  \label{fig:select_edges_good}
       \includegraphics[scale=0.40]{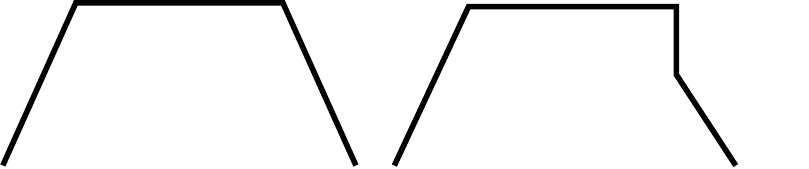}
    }        
  \subfigure[]
    {  \label{fig:select_edges_bad}
       \includegraphics[scale=0.40]{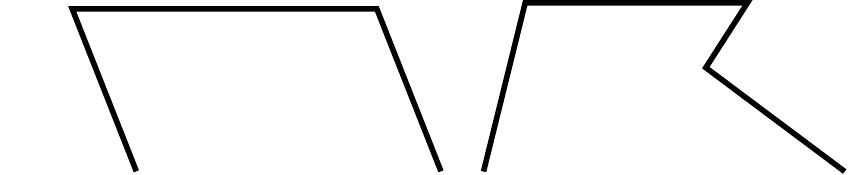}
    }    
   \caption{(a) Desired corridor configurations where our algorithm can select all the ground edges.  
   (b) Unsatisfactory configurations because of too much overlap horizontally. Our algorithm might miss some ground edges.}
\end{figure}

The second constraint is that edges should not overlap with each other beyond a threshold
in image horizontal direction shown in Figure \ref{fig:select_edges_good}.
This is true for most cases in the real world. In the latter experiments, we find that 
even for the unsatisfactory configurations in Figure \ref{fig:select_edges_bad}, 
our algorithm can select most of the ground edges. We can denote this constraint as:
\begin{equation}
\label{eq:edge constraint 2}
I_{ovlp}=\lbrace S\colon \forall e_i, e_j \in S, O(e_i,e_j)< \delta_{ovlp} \rbrace
\end{equation}
where $O$ is horizontal overlapping length between two edges.

Similarly, we want to maximize the covering of edges in image $x$ direction. So the score
function is defined as:
\begin{equation}
\label{eq:edge cost function}
F\colon \lbrace S\rightarrow \mathbb{R}, \ F= C(S) \rbrace
\end{equation}
where $C$ is the horizontal covering length of edge sets $S$.

With the defined score function $F$ and constraints $I=I_{close} \cap I_{ovlp}$, the problem changes to a
submodular set optimization. We adopt a greedy algorithm \cite{krause2012submodular} to select the edges in sequence. 
We initially start with an empty set of edges $S$, then iteratively add edges by:
\begin{equation}
\label{eq:greedy}
S\leftarrow S \cup \lbrace \argmax_{e \notin S: S \cup{ \lbrace e \rbrace} \in I} \bigtriangleup(e\mid S)   \rbrace
\end{equation}
until there is no feasible edges. $\bigtriangleup(e\mid S)$ is the marginal gain of adding edge $e$ into set $S$. 
Details and proof of submodularity and optimality are in the appendix.

After getting the edge set $S$, some post processing steps are required for example
removing tiny edges and merging adjacent edges into a longer one similar to \cite{hedau2009recovering}.

\subsection{Pop-up World from an Arbitrary Pose}
\label{sec:pop_up_pose}
%This section will pop up 3D plane model based on the above selected ground edges $S$. 
%We start by briefly introducing the state representation then provide details of pop up process.
%Instead of assuming a fixed camera pose in \cite{chao2016pop}, we modify to pop up using an arbitrary pose.

%A homogeneous point $p=(x,y,z,1)^\top$ lies on plane iff $\pi^\top p=0$.
\textbf{Notations.}
We use subscript $w$ to represent global \textit{world} frame and $c$ to denote local \textit{camera} frame. 
$gnd$ is short for ground plane.
A plane can be represented as a homogeneous vector $\pmb{\pi}=(\pi_1,\pi_2,\pi_3,\pi_4)^\top = (\mathbf{n}^\top,d)^\top$,
where $\mathbf{n}$ is the plane normal vector, and $d$ is its distance to the origin \cite{hartley2003multiple} \cite{kaess2015simultaneous}. The camera pose is represented by the 3D Euclidean transformation matrix $\text{T}_{w,c} \in \mathbf{SE}(3)$ from local to global frame. 
Then a local point $\mathbf{p}_c$ can be transformed to global frame by: $\mathbf{p}_w=T_{w,c}\mathbf{p}_c$, and a local
plane $\pmb{\pi}_c$ is transformed to global frame by:
\begin{equation}
\label{eq:plane transform}
\pmb{\pi}_{w} = \text{T}_{w,c}^{-\top} \pmb{\pi}_{c}
\end{equation}

\subsubsection{Create 3D model}
For each image pixel $\mathbf{u}\in \mathbb{R}^3$ (homogeneous form) belonging to a certain local plane $\pmb{\pi}_{c}$, 
the corresponding 3D pop-up point $\mathbf{p}_c$ is the intersection of backprojected ray $\text{K}^{-1}\mathbf{u}$ with plane $\pmb{\pi}_c$:
\begin{equation}
\label{eq:projection equation}
\mathbf{p}_c=\frac{-d_c}{\mathbf{n}_c^\top(\text{K}^{-1}\mathbf{u})} \text{K}^{-1}\mathbf{u}
\end{equation}
where $\mathbf{K}$ is calibration matrix.

Then we show how to compute the plane equation $\pmb{\pi}_c$. Our world frame is built on the ground plane represented
by $\pmb{\pi}_{gnd,w}=\left(0,0,1,0 \right)^\top$. Suppose a ground edge's boundary pixels are $\mathbf{u}_0, \mathbf{u}_1$,
their 3D point $\mathbf{p}_{c0}, \mathbf{p}_{c1}$ can be computed by Equation \eqref{eq:plane transform} \eqref{eq:projection equation}. Using the assumption that wall is vertical to the ground, we can compute the wall plane normal by:
\begin{equation}
\label{eq:wall normal}
\mathbf{n}_{wall,c}= \mathbf{n}_{gnd,c} \times \left(\mathbf{p}_{c1}-\mathbf{p}_{c0}\right)
\end{equation}
We can further compute $d_{wall,c}$ using the constraints that two points $\mathbf{p}_{c0}, \mathbf{p}_{c1}$ lying on the wall.

\subsubsection{Camera pose estimation}
The camera pose $\text{T}_{w,c}$ could be provided from other sensors or state estimation methods. 
Here, we show a single image attitude estimation method which could be used at the
SLAM initialization stage. For a Manhattan environment, there are three orthogonal dominant directions 
$\mathbf{e}_1=\left(1,0,0 \right)^\top,\mathbf{e}_2=\left(0,1,0 \right)^\top,\mathbf{e}_3=\left(0,0,1 \right)^\top$ 
corresponding to three vanishing points $\mathbf{v}_1,\mathbf{v}_2,\mathbf{v}_3 \in \mathbb{R}^3$ in
homogeneous coordinate. If the camera rotation matrix is $\mathbf{R}_{w,c}\in \mathbb{R}^{3\times3}$, then $\mathbf{v}_i$ 
can be computed by \cite{hedau2009recovering} \cite{rother2002new}:
\begin{equation}
\label{eq:vanish point}
\mathbf{v}_i=\mathbf{K} \mathbf{R}_{w,c}^\top \mathbf{e}_i, \quad i \in \left\lbrace 1,2,3 \right\rbrace
\end{equation}

With three constraints of Equation \eqref{eq:vanish point}, we can recover the 3 DoF rotation $\mathbf{R}_{w,c}$. 
%To obtain an absolute scale, we assume the camera is $1$m above the ground.

%They are computed by clustering the detected 2D line segments into three dominant clusters\cite{rother2002new} \cite{hedau2009recovering}. 

\section{Pop-up Plane Slam}
\label{sec:plane slam}
This section introduces the \textit{Pop-up Plane SLAM} using monocular images.
Plane SLAM has recently been addressed by Kaess \cite{kaess2015simultaneous} with a RGB-D sensor,
here we extend it to the monocular case based on the pop-up plane model.
%with infinite planes as landmarks has been addressed 

\subsection{Planar SLAM Formulation}
\begin{figure}[thpb]
\vspace{2.0 mm}
  \centering
   \includegraphics[scale=0.53]{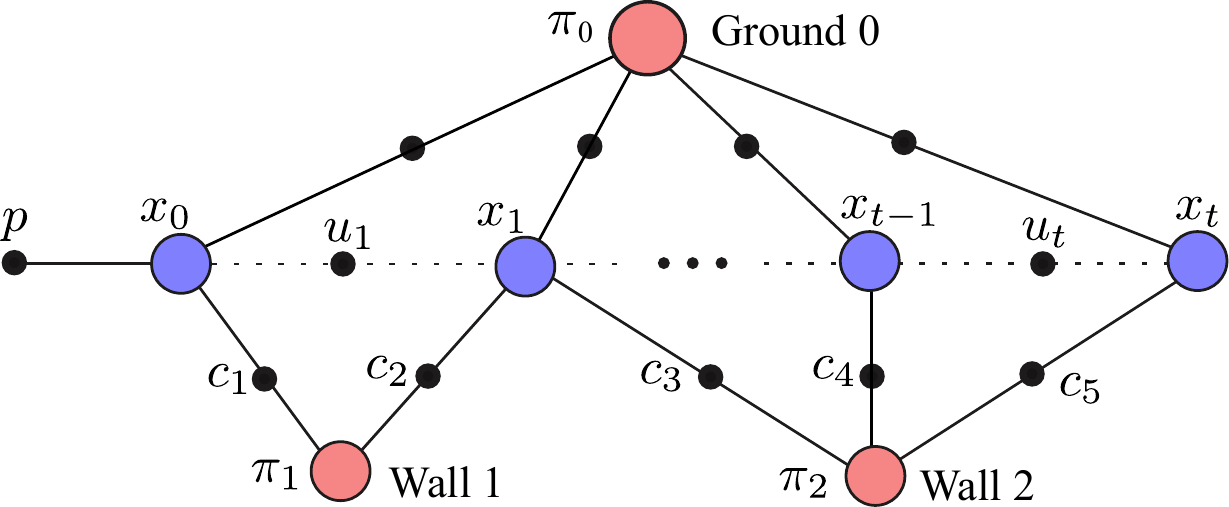}  %width=0.5
   \caption{Plane SLAM factor graph. Variable nodes include camera pose $x$, plane landmark $\pi$. Factor
   nodes are odometry measurements $u$ and plane measurements $c$. The latter come from the single image pop-up model.
   Each plane node $\pi$ also has a label of either ground or wall.}
   \label{fig:slam_graph}
\end{figure}

The factor graph of planar SLAM is shown in Figure \ref{fig:slam_graph}. We need to estimate
the 6 DoF camera poses $x_0,...,x_t$ and plane landmarks $\pmb{\pi}_0,...,\pmb{\pi}_n$ 
using the plane measurements $c_0,...,c_m$, odometry measurements $u_1,...,u_t$ and initial 
pose constraint $p$. Note that, our plane landmark also has a label being either ground or wall. 
The ground plane landmark $\pmb{\pi}_0$ is connected to all pose nodes. 

The homogeneous plane representation $\pmb{\pi}=(\mathbf{n}^\top,d)^\top$ is over-parametrized and therefore 
the information matrix of SLAM is singular and not suitable for Gauss-Newton solver and incremental 
solvers such as iSAM \cite{kaess2008isam}. We utilize the minimal plane representation in \cite{kaess2015simultaneous}
to represent planes as a unit quaternion $\mathbf{q}=(q_1,q_2,q_3,q_4)^\top \in \mathbb{R}^4 $ st. $\|q\|=1$.
We can therefore use Lie algebra and exponential map to do plane updates during optimization. 

%The the over-parameterized homogeneous plane representation $\pmb{\pi}=(\mathbf{n}^\top,d)^\top$ in Section \ref{sec:pop_up_pose}
%For the SLAM optimization, instead of using the over-parameterized homogeneous plane representation 
%$\pmb{\pi}=(\mathbf{n}^\top,d)^\top$ in Section \ref{sec:pop_up_pose}, 
%we use the minimal plane representation as a quaternion $\mathbf{q}=(q_1,q_2,q_3,q_4)^\top \in \mathbb{P}^3$,
%which enforces $\pmb{\pi}$ to lie on the unit sphere $\Vert \pmb{\pi} \Vert _2 =1$ \cite{kaess2015simultaneous}. 
%We can therefore use Lie algebra and exponential map to do plane updates during optimization. 
%iSAM \cite{kaess2008isam} library is used to perform efficient optimization.

\subsection{Plane Measurement}
\label{sec:plane measurement}
Most plane SLAM \cite{kaess2015simultaneous, taguchi2013point} uses RGB-D sensor to get
plane measurements $c$ from the point cloud segmentation. In our system, plane measurements $c$ 
come from the pop-up plane model in Section \ref{sec:pop_up_pose}. Note that the pop-up 
process depends on the camera pose, more specifically rotation and height because camera $x,y$ position
does not affect local plane measurements. So we need to re-pop up the 3D plane model and update plane 
measurement $c$ after camera poses are optimized by plane SLAM. This step is fast with simple matrix 
operation explained in Section \ref{sec:pop_up_pose}. It takes less than 1ms to update 
hundred's plane measurement.
%analysis, maybe in separate section  not height? maybe not, because lsd can have height

\subsection{Data Association}
We use the following three geometry information for plane matching: the difference between plane normals, 
plane distance to each other and projection overlapping between planes. The plane's bounding polygon for 
projection comes from the pop-up process. Outlier matches are first removed by
thresholds of the three metrics. Then the best match is selected based on a weighted sum of them.

%Geometry information is used to match the planes. The matching score between two planes is a weighted sum of three
%terms: Thresholds of these three metrics 

%then project planes onto each other to compute the overlapping in Figure \ref{fig:association}.
% We finally compute a weighted sum of these three 
%etrics as the cost and select the matches with the lowest cost.

%instead of visual features to improve the robustness in low-texture environments. 
%For point based SLAM, we usually use visual feature descriptor to match the points,
%but it is difficult in texture-less environment such as Figure \ref{fig:method overview}.

\begin{figure}[thpb]
  \centering
   \includegraphics[scale=0.65]{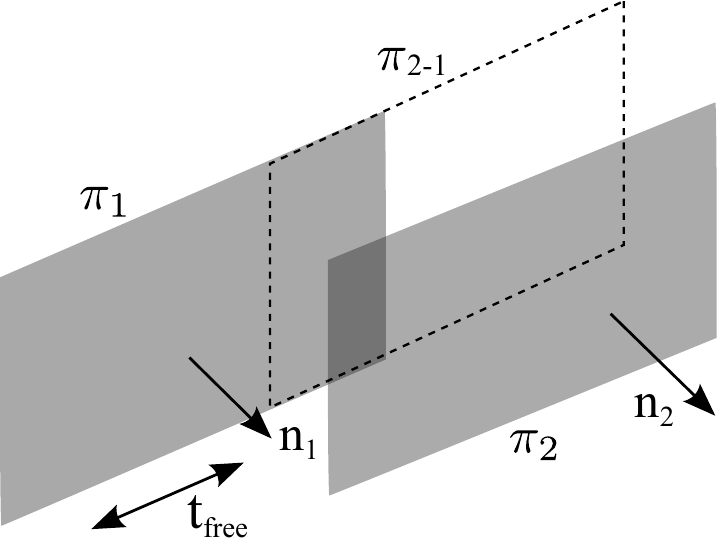}
   \caption{Data association and unconstrained situations. $\pi_1, \pi_2$ are two planes with normals $n_1,n_2$ respectively.
   $\pi_{2\_1}$ is the projected plane from $\pi_2$ onto $\pi_1$, used for data association. 
   In this example, $n_1$ and $n_2$ are parallel so there is an unconstrained direction along $t_{free}$.}
   \label{fig:association}
\end{figure}

\subsection{Loop Closure}
\label{sec:Loop Closure}
\begin{figure}[thpb]
  \centering
   \includegraphics[scale=0.47]{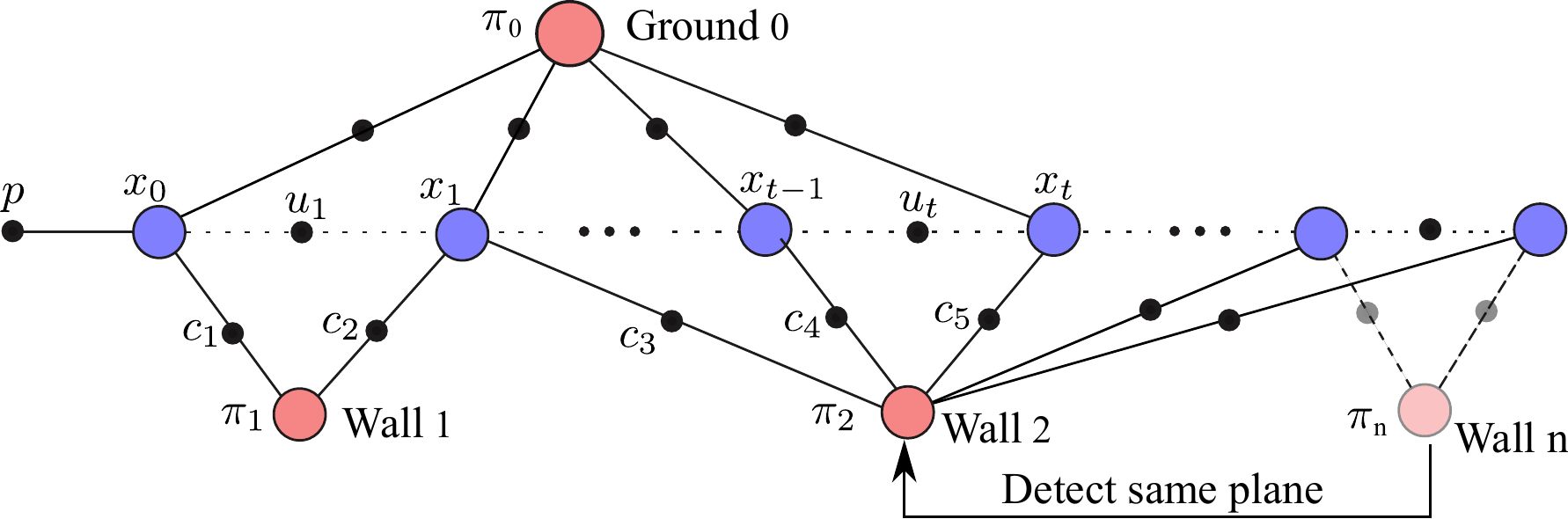}
   \caption{Plane SLAM loop closure. After detecting a loop closure, we shift all the factors of plane $\pi_n$ 
   to $\pi_2$ and remove $\pi_n$ from factor graph.}
   \label{fig:loop_closing}
\end{figure}

%We realize the plane SLAM loop closure in this section while most existing plane SLAM systems do not
%\cite{kaess2015simultaneous} \cite{concha2015dpptam}. 
We adopt a bag of words (BoW) place recognition method \cite{galvez2012bags} for loop detection. 
Each frame is represented as a vector of visual worlds computed by ORB descriptors so as to 
calculate the similarity score between two frames. Once a loop closure frame is detected, we 
search all the plane pairs in the two frames and find the plane pairs with smallest image space distance. 
We also tested to keep BoW visual words for each plane, but it is not robust especially in texture-less images.
Different from point landmarks, plane landmarks have different appearance and size in different views. So we may 
recognize the same planes after the landmark has been created and observed for sometime.
So after detecting, for example, $\pi_n$ and $\pi_2$ as the same plane in Figure \ref{fig:loop_closing}, 
we shift all the factors of plane $\pi_n$ to the other plane $\pi_2$, and remove the landmark $\pi_n$ from
factor graph.

\section{Point-Plane SLAM Fusion}
\label{sec:point plane SLAM}
Compared to point based SLAM, planar SLAM usually contains much fewer landmarks so it becomes
easily unconstrained. For example in a long corridor in Figure \ref{fig:association} where left
and right walls are parallel, there is a free unconstrained direction $t_{free}$ along corridor
if there is no other plane constraints. We solve this problem by incorporating with point based SLAM, 
specifically LSD SLAM \cite{engel2014lsd}, to provide photometric odometry constraints along the
free direction. We propose the two following combinations.

\subsection{Depth Enhanced LSD SLAM}
\label{sec:enhanced LSD}
\begin{figure}[thpb]
\vspace{2.0 mm}
  \centering
   \includegraphics[scale=0.48]{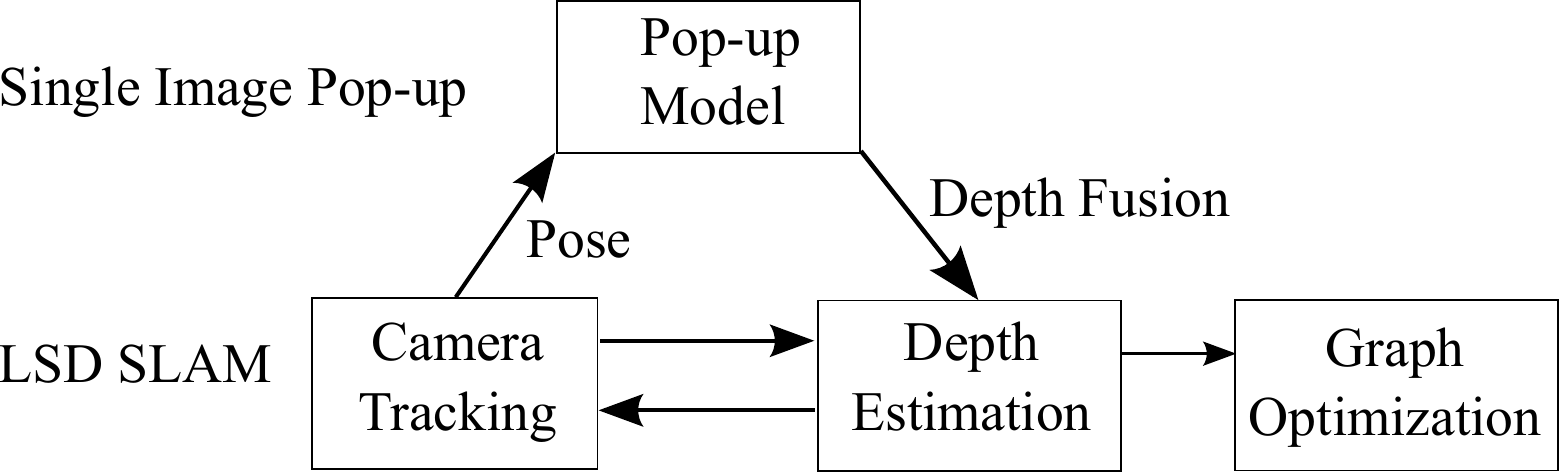}
   \caption{\textit{Depth Enhanced LSD SLAM} algorithm that integrates depth estimates from the pop-up model.}
   \label{fig:enhanced_lsd}
\end{figure}

This section shows that scene layout understanding could boost the performance 
of traditional SLAM. LSD SLAM has three main threads: camera tracking, 
depth estimation and global optimization in Figure \ref{fig:enhanced_lsd}. The core part is depth estimation, 
determining the quality of other modules. In LSD SLAM, when a new depth map of a keyframe
is created, it propagates some pixels' depth from the previous keyframe if it is available. 
Then the depth map is continuously updated by new frames using multiple-view stereo (MVS). 
Since our single image pop-up model in Section \ref{sec:single image pop} provides each pixel's depth estimation, 
we integrate its depth into LSD depth map in the following way:

(1) If a pixel has no propagated depth or the variance of the LSD SLAM depth exceeds a threshold, 
we directly use pop-up model depth.

(2) Otherwise, if a pixel has a propagated depth $d_l$ with variance $\sigma_l^2$ from LSD, we fuse it with the pop-up depth
$d_p$ of variance $\sigma_p^2$ using the filtering approach \cite{engel2013semi}:

\begin{equation}
\label{eq:depth_fusion}
\mathcal{N} \left(  \frac{\sigma_l^2 d_p+\sigma_p^2 d_l }{\sigma_l^2+\sigma_p^2}, \frac{\sigma_l^2 \sigma_p^2}{\sigma_l^2+\sigma_p^2}  \right)
\end{equation}
$\sigma_p^2$ could be computed by error propagation rule during the pop-up process. In Section 
\ref{sec:pop_up_pose}, the pixel uncertainty of $\mathbf{u}$ can be modeled as bi-dimensional standard Gaussians $\Sigma_u$. 
If the Jacobian of $\mathbf{p}_c$ \textit{wrt.} $\mathbf{u}$ is $J_u$ from Equation \eqref{eq:projection equation},
then the 3D point's covariance $\Sigma_{p_c}=J_u \Sigma_u J_u ^\top$. We find that depth uncertainty $\sigma_p^2$ is proportional to the depth square, namely $\sigma_p^2 \propto d_p^2$.

%If we model the pixel $\mathbf{u}$  uncertainty $\Sigma_u$ as bi-dimensional standard Gaussians, then we can derive the Jacobian of $\mathbf{p}_c$ \textit{wrt.} $\mathbf{u}$ as $J_u$ and compute 3D point's covariance $\Sigma_p=J_u \Sigma_u J_u ^\top$. It is proportional to the depth square, namely $\sigma_p^2 \propto d_p^2$. 

%For simplicity, $\sigma_p^2$ is set using the assumption that it is proportional to the depth square, namely $\sigma_p^2 \propto d_p^2$, which are found to work well in experiments. 
%We tune the weights so that two variance $\sigma_l^2, \sigma_p^2$ are around the same order of magnitude.

%We set the pop-up variance $\sigma_p^2$ based on the assumption that the variance is proportional to
%the fourth power of depth value \cite{engel2013semi}, namely $\sigma_p^2 \propto d_p^4$. 

Depth fusion could greatly increase the depth estimation quality of LSD SLAM especially at the initial frame where LSD SLAM just
randomly initializes the depth and at the low parallax scenes where MVS depth triangulation has low quality. This
is also demonstrated in the latter experiments.

\subsection{LSD Pop-up SLAM}
\label{sec:Pop LSD SLAM}
There has been some work jointly using point and plane as landmarks in one SLAM framework \cite{taguchi2013point} using RGB-D sensors. 
Currently, we propose a simple version of it to run two stages of SLAM methods. 
The first stage is \textit{Depth Enhanced LSD SLAM} in Section \ref{sec:enhanced LSD}. We then use its pose output 
as odometry constraints to run a plane SLAM in Section \ref{sec:plane slam}.
The frame-to-frame odometry tracking based on photometric error minimization could provide constraints along the 
unconstrained direction in plane SLAM and can also capture the detailed fine movements, demonstrated in the 
latter experiments.

\begin{figure}[thpb]
  \centering
   \includegraphics[scale=0.286]{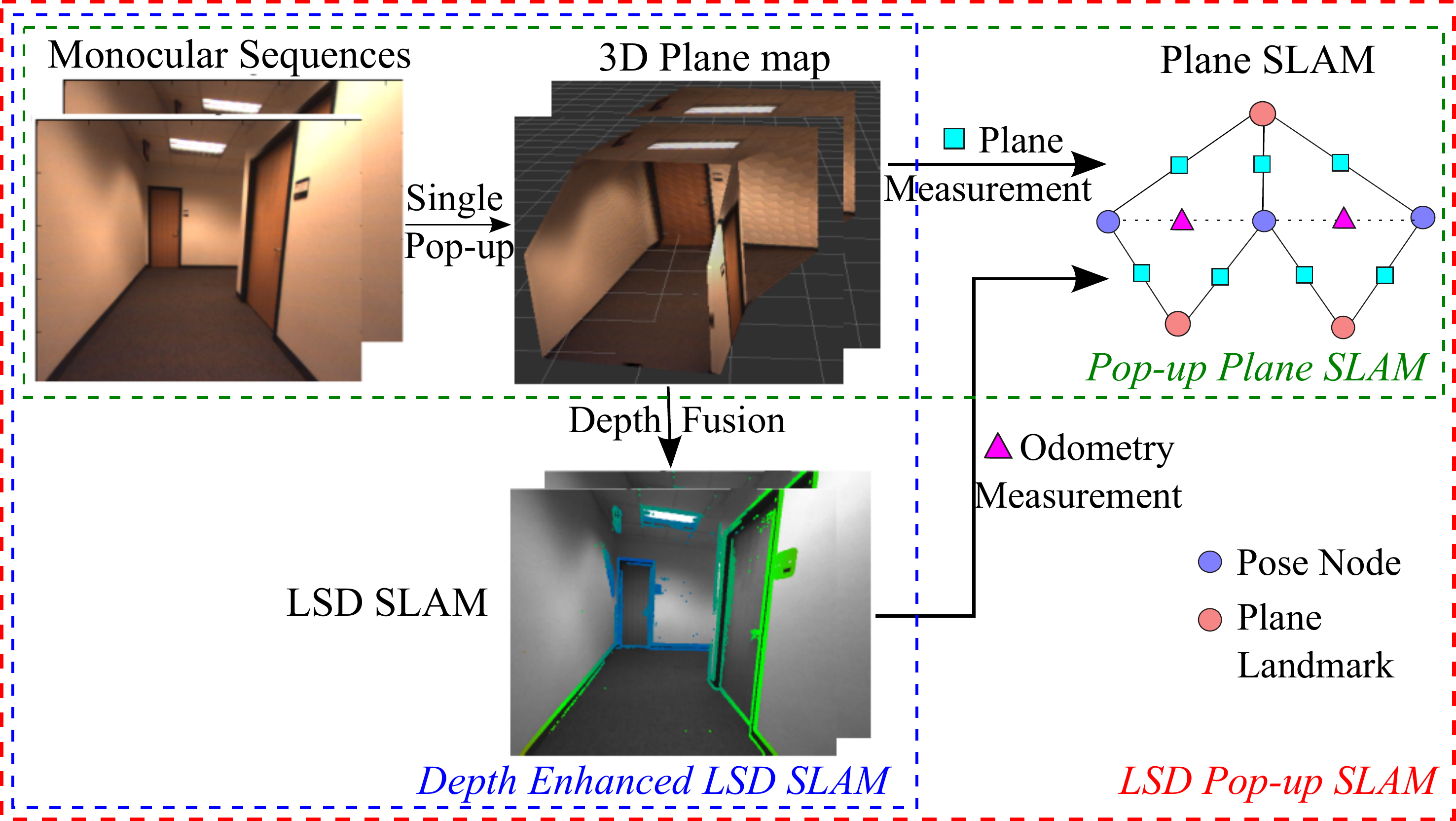}
   \caption{ Relationship between three proposed SLAM methods: (1) \textit{Pop-up Plane SLAM} uses plane measurements
   from the single image pop-up model. (2) \textit{Depth Enhanced LSD SLAM} is LSD SLAM with depth fusion from the pop-up model.
   (3) \textit{LSD Pop-up SLAM} is plane SLAM with additional odometry measurements from \textit{Depth Enhanced LSD SLAM}.}
   \label{fig:slam_overview}
\end{figure}

Figure \ref{fig:slam_overview} shows the relationship of the three SLAM methods in this paper. The blue dashed box 
is the improved LSD SLAM: \textit{Depth Enhanced LSD SLAM}. The green and red box show two kinds of plane SLAM. 
The difference is that \textit{LSD Pop-up SLAM} in this section has additional odometry measurement while 
\textit{Pop-up Plane SLAM} does not have and usually uses a constant velocity assumption.

\section{Experiments and results}
\label{sec:experiments}
We test our SLAM approaches on both the public TUM dataset \cite{sturm2012benchmark} and two collected 
corridor datasets to evaluate the accuracy and computational cost. Results can also be found in the supplemental videos. We compare the state estimation and 3D reconstruction quality with two state-of-art point based monocular SLAM approaches: LSD SLAM \cite{engel2014lsd} and ORB SLAM \cite{mur2015orb}.

%Some plane or superpixel dense mapping algorithms \cite{furukawa2010accurate} \cite{concha2015dpptam} 
%depend on those point based SLAM or bundle adjustment to provide state estimation to provide state estimation 
%so we didn't need to compare with them.
%PTAM \cite{klein2007parallel} or LSD SLAM

%to demonstrate dense mapping in real time especially in texture-less environments.
\subsection{TUM SLAM dataset}
We choose the TUM \textit{fr3/structure\_notexture\_far} dataset in Figure \ref{fig:method overview}, 
which is a challenging environment composed of five white walls and a ground plane. 
We only use RGB images for experiments and use the depth images for evaluation.

%We use the constant velocity motion model if there is no other odometry fed in.
%using the single  image pose estimation in Section \ref{sec:pop_up_pose}

%and all the latter pose and map estimates are based on that scale.

\subsubsection{Qualitative Comparison}
Unfortunately, neither LSD nor ORB SLAM work in this environments because there are only few features 
and high gradient pixels. 

For the \textit{Pop-up Plane SLAM} in Section \ref{sec:plane slam}, we use the ground truth pose for 
initialization and a constant velocity motion assumption as odometry measurements. Since the initial 
truth height is provided, the pop-up model has an absolute scale. Therefore, we can directly compare the 
pose and map estimates with ground truth without any scaling. The constructed 3D map is 
shown in Figure \ref{fig:method overview}.
%Our \textit{Pop-up Plane SLAM} algorithm can still work and construct 3D map 

%Since LSD SLAM doesn't work, the proposed two versions of point plane fusion algorithm:
%\textit{Depth Enhanced LSD SLAM} and \textit{LSD Pop-up SLAM} in Section \ref{sec:point plane SLAM} 
%also don't work.

%There is also semantic label on the plane landmark of either ground or a certain wall.
%The walls are vertical relative to each other and also to the ground
%no shown on the 3D map. Each map point has a label belonging to either ground plane or a certain wall plane.
%Also, the 3D map is in absolute scale, which will be compared with ground truth depth latter.

\subsubsection{Quantitative Comparison}
\begin{figure}[thpb]
\vspace{2.0 mm}
  \centering
   \includegraphics[scale=0.13]{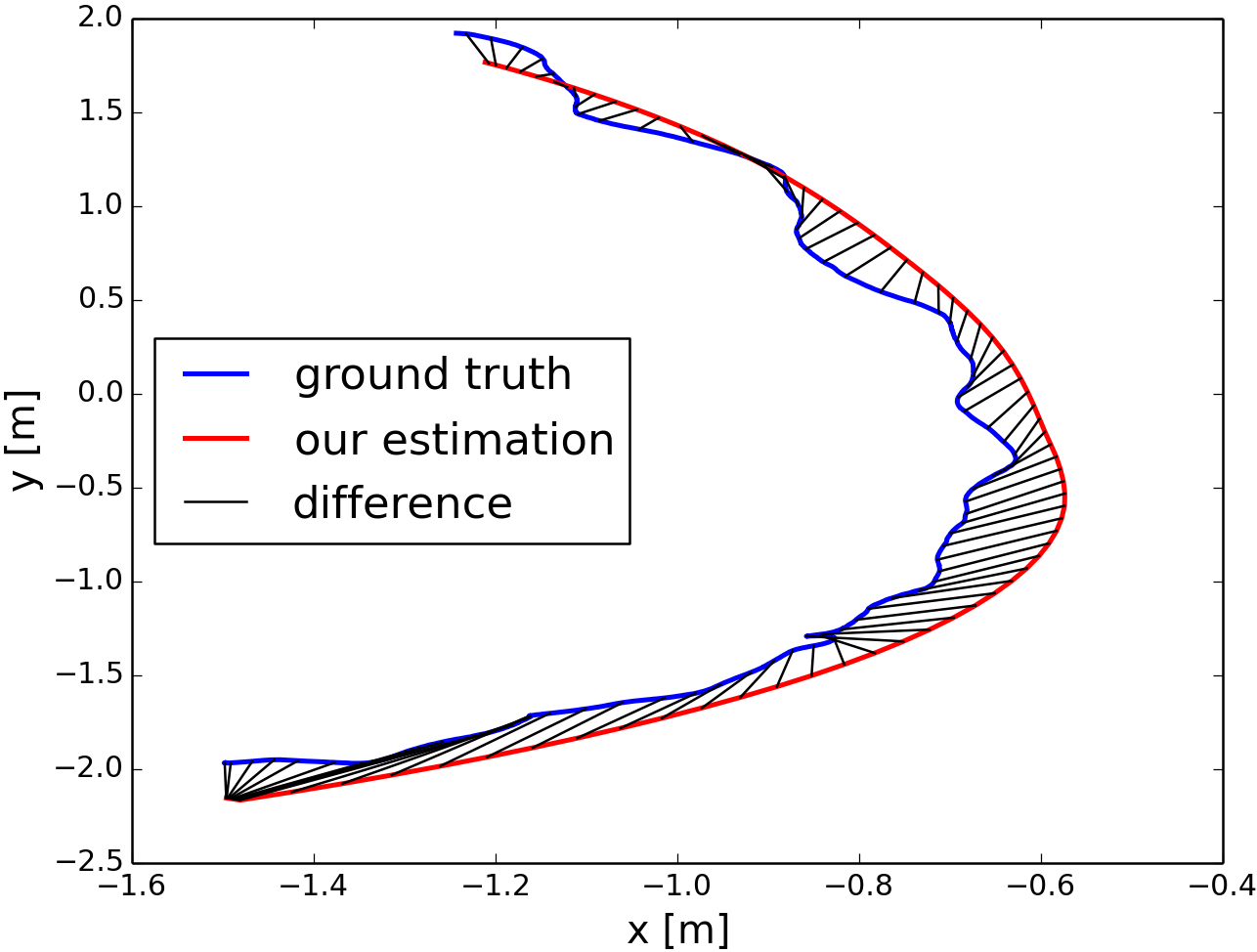}
   \caption{Absolute trajectory estimation using \textit{Pop-up Plane SLAM} on TUM \textit{fr3/str\_notex\_far} dataset.
   The positioning error is 3.9\%, while LSD SLAM and ORB SLAM both fail.}
   \label{fig:tum trajectory}
\end{figure}

The absolute trajectory estimate is shown in Figure \ref{fig:tum trajectory}. This dataset
has a total length of $4.58$m and our mean positioning error is $0.18 \pm 0.07$m, with endpoint error $0.10$m.
From Figure \ref{fig:tum trajectory}, our algorithm captures the overall 
movement but not the small jerk movement in the middle. This is mainly 
due to the fact that there are only few plane landmarks in SLAM. In addition, \textit{Pop-up Plane SLAM} 
does not have frame-to-frame odometry tracking to capture the detailed movement, which is commonly used in point 
based SLAM. In the latter experiments, we show that after getting odometry 
measurements, state estimation of \textit{LSD Pop-up SLAM} improves greatly.

\iffalse
\begin{table}[h]
\caption{Positioning error on TUM \textit{fr3/str\_notex\_far} dataset}
\label{table:Tum traj compare}
 \begin{center}
     \begin{tabular}{c c c c c c c} 
      \toprule      
      \multirow{2}{*}{Name}	 &\multirow{2}{*}{LSD SLAM}     &\multirow{2}{*}{ORB SLAM}       &Pop-up    \\  
       				          &           &                  &Plane SLAM   \\  \midrule
      Positioning error      &Fail     &Fail      &3.9\%  \\ \bottomrule
      \end{tabular}
 \end{center}
\end{table}
\fi

\begin{figure}[thpb]
  \centering
   \includegraphics[scale=0.40]{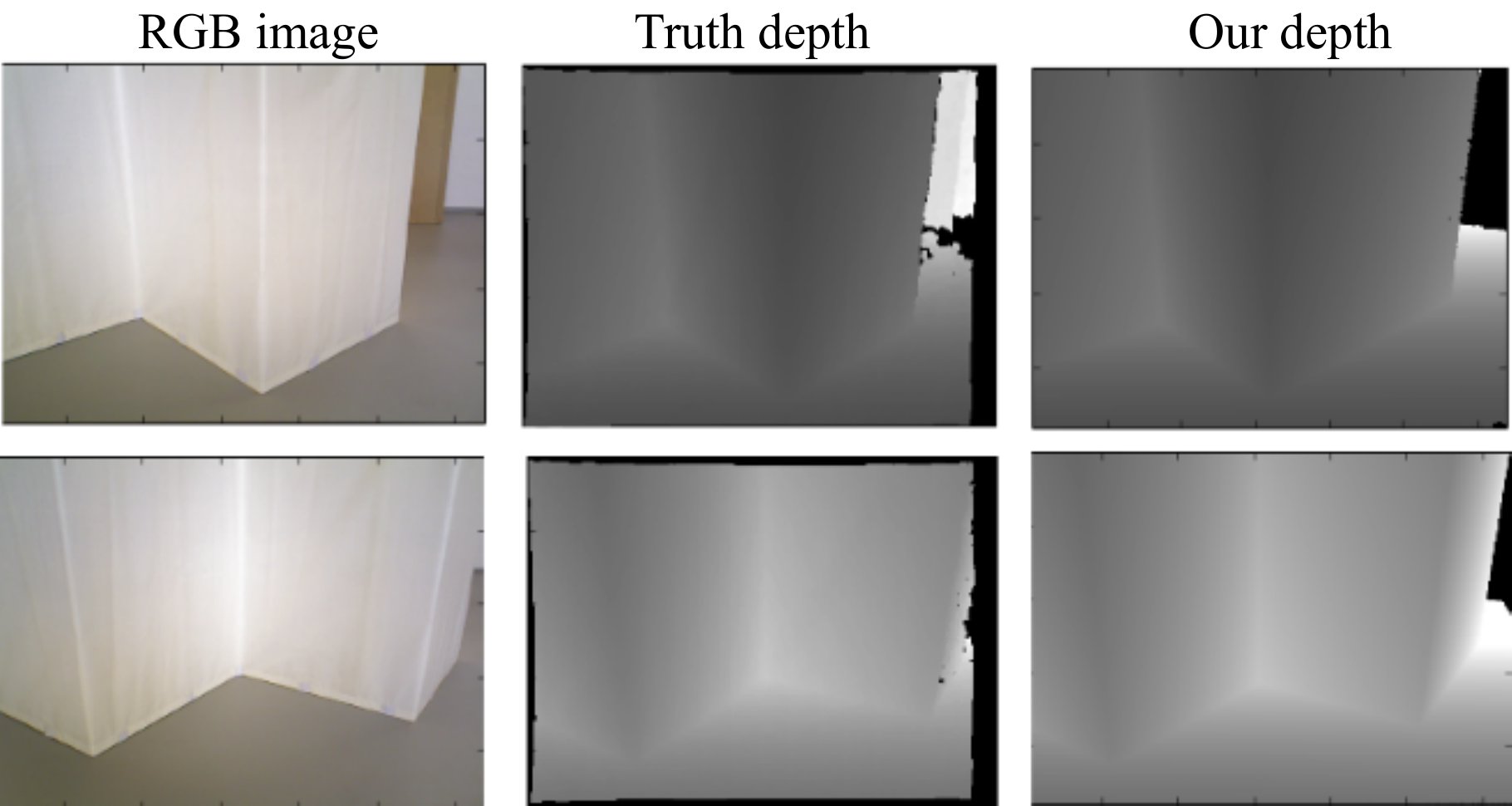}
   \caption{Depth reconstruction comparison on TUM dataset.}
   \label{fig:Depth comapre}
\end{figure}

To evaluate mapping quality, we use provided depth maps to compute the ground truth plane position 
by point cloud plane segmentation using the PCL RANSAC algorithm. The plane normal error is only $2.8 \degree$ as shown in Table \ref{table:Tum depth compare}. We then re-project the 3D plane model onto images to get each pixel's depth estimates. The evaluation result is shown in
Figure \ref{fig:Depth comapre} and Table \ref{table:Tum depth compare}. The mean pixel depth error is $6.2$ cm and 
$86.8\%$ of the pixels' depth error is within $0.1$m.

%We then project each frame's pop-up cloud onto the optimized SLAM planes to get a dense 3D model then 
%re-project it onto images using calibration matrix to get each pixel's depth estimates.

\begin{table}[h]
\caption{3D Reconstruction Evaluation on TUM dataset.}
\label{table:Tum depth compare}
 \begin{center}
     \begin{tabular}{c c c c c c c} 
      \toprule
      {}    				  &Plane normal error    &Depth error     &Depth error $<$ 0.1m   \\  \midrule
      Value               &2.83$\degree$    		&6.2 cm         &86.8\%    \\ \bottomrule
      \end{tabular}
 \end{center}
\end{table}

\subsection{Large Indoor Environment}
In this section, we present experimental results using a hand-held monocular camera with a resolution of
640 $\times$ 480 in two large low-texture corridor environments. The camera has a large field of view ($\sim90\degree$)
which LSD and ORB SLAM typically prefer. Since we do not have ground truth depth or pose, we 
only evaluate the loop closure error and qualitative map reconstruction. The pose initialization uses the single 
image rotation estimation in Section \ref{sec:pop_up_pose} with an assumed height of 1m.

\subsubsection{Corridor dataset I}
\begin{figure}[thpb]
\vspace{1 mm}
  \centering
   \includegraphics[scale=0.30]{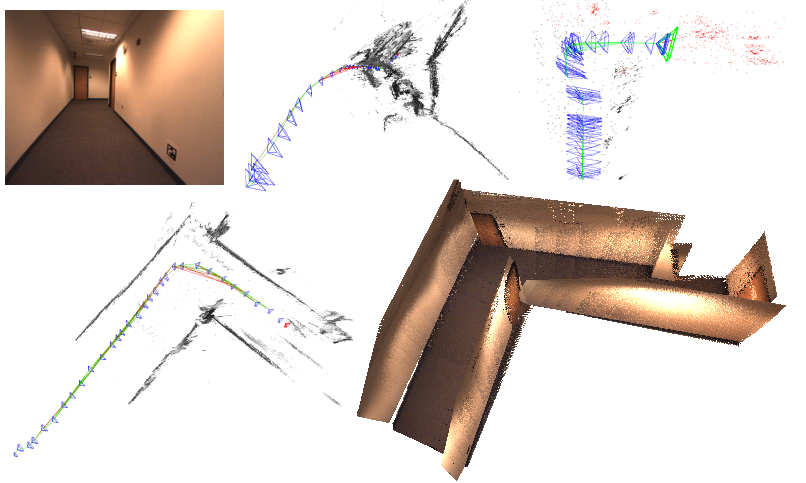}
   \caption{Corridor dataset I. (top) From left to right: sample frame, LSD SLAM result, ORB SLAM result. 
   (bottom) Our \textit{Enhanced LSD SLAM} in Section 
   \ref{sec:enhanced LSD}, \textit{LSD Pop-up SLAM} result in Section \ref{sec:Pop LSD SLAM}.}
   \label{fig:nsh 1}
\end{figure}

The first dataset is shown in Figure \ref{fig:nsh 1}. LSD SLAM, top center, does not perform well.
The best result for ORB SLAM is shown in the top right. Through the tests, 
we find that even using the same set of parameters, ORB SLAM often cannot initialize the map 
and fails to track cameras. The randomness results from the RANSAC 
map initialization of ORB SLAM \cite{mur2015orb}. 
%It may initialize very different maps especially when there are few features in this environment. 
%There is no public implementation for semidense mapping of ORB SLAM[], but even it works, it can only model 
%high gradient edges while large remaining planar surfaces cannot be mapped.

Since actual long corridors are easily under-constrained, \textit{Pop-up Plane SLAM} with no actual odometry measurement 
does not work well. We only provide results for the two other SLAM methods introduced in Section \ref{sec:point plane SLAM}.
\textit{Depth Enhanced LSD SLAM} generates a much better map as shown in the bottom left 
of Figure \ref{fig:nsh 1} compared to the original LSD SLAM. Though it is a semi-dense map, we can clearly see 
the passageway and turning. Based on that, the \textit{LSD Pop-up SLAM} generates a dense 3D model with distinct 
doors and pillars.
%Also, note that our map is has a absolute scale, which be shown more clearly in next dataset.

\subsubsection{Corridor dataset II}
\begin{figure}[thpb]
\vspace{2.0 mm}
  \centering
   \includegraphics[scale=0.33]{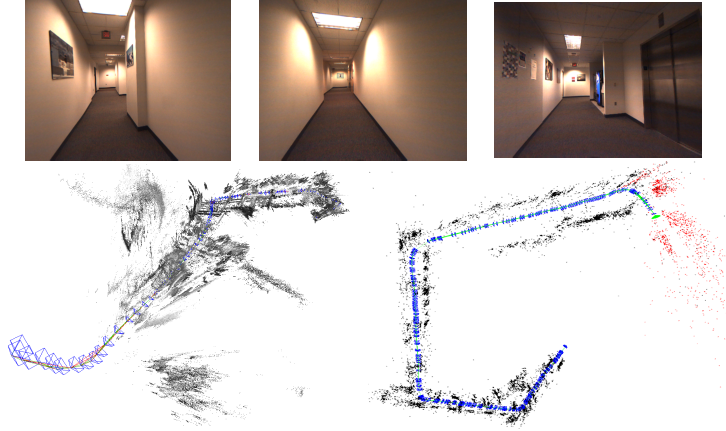}
   \caption{Corridor dataset II with loop closure. (top) Sample frames in  
   the dataset. (bottom): LSD SLAM result, ORB SLAM result.}
   \label{fig:nsh 2 1}
\end{figure}

The second dataset is a $60$m square corridor containing a large loop shown in Figure \ref{fig:nsh 2 1}. 
ORB SLAM generates a better map than LSD SLAM, but it does not start tracking until it comes to a large open 
space with more features and enough parallax.
%LSD SLAM and ORB SLAM results are in the bottom row. 

\begin{figure}[thpb]
\vspace{2.5 mm}
  \centering
   \includegraphics[scale=0.35]{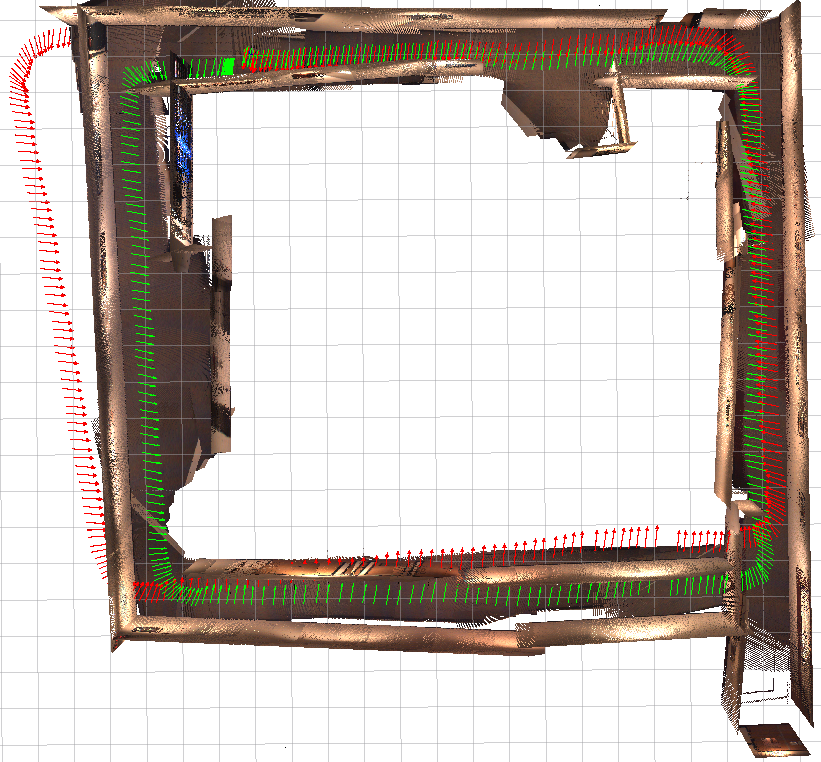}
   \caption{Corridor dataset II with loop closure (continued). Red line: \textit{Enhanced 
   LSD SLAM} result in Section \ref{sec:enhanced LSD}, green line: \textit{LSD Pop-up SLAM} result in Section \ref{sec:Pop LSD SLAM}.
   Loop closure happens around the top left corner. The grid dimension is 1m. Loop closure positioning error is
   $0.67\%$.}
   \label{fig:nsh 2 2}
\end{figure}

The result of our algorithms is shown in Figure \ref{fig:nsh 2 2} where the red line is \textit{Enhanced 
LSD SLAM} and green line is \textit{LSD Pop-up SLAM}. With the automatic loop closure detection, 
the \textit{LSD Pop-up SLAM} generates the best 3D map and smallest loop closure error. 
The grid dimension is $1\times 1 m^2$ in Figure \ref{fig:nsh 2 2} and the loop closure 
positioning error is $0.4$m of the total $60$m trajectory.

%and our algorithm keeps the absolute scale using the initializing height of $1m$.

\subsection{Runtime Analysis}
Finally, we provide the computation analysis of the Corridor dataset II in Table \ref{table:Tum time}. 
All timings are measured on CPU (Intel i7, 4.0 GHz) and GPU (only for CNN). Most of the code is 
implemented in C++. Currently, the CNN segmentation, edge detection, and selection consumes $30$ms. 
Note that compared to CNN model in \cite{chao2016pop}, we change the fully connected layers from 4096 to 2048
to reduce the segmentation time by half without affecting the accuracy too much. The iSAM 
incremental update takes $17.43$ms while batch optimization takes $45.35$ms. Therefore we only use batch 
optimization to re-factor the information matrix when a loop closure is detected. In all, our 
plane SLAM algorithm can run in real time at over $20$Hz using 
single-threaded programming.
%and could be increased using muti-threaded programming for segmentation and edge detection
%Actually, we did not find much performance difference between them if there is no loop closure.

We also note that unlike point landmarks, a plane landmark can be observed in many adjacent frames, 
so we actually do not need to pop up planes for each frame. Thus in all the above pop-up experiments, 
we process the images at $3$Hz (every 10 images), which we find is enough to capture all the planes.
This is similar to the keyframe techniques used in many point-based SLAM algorithms.

\begin{table}[h]
\vspace{2.0 mm}
\caption{SLAM Statistics and Time Analysis on Corridor Dataset II.}
\label{table:Tum time}
 \begin{center}
     \begin{tabular}{c c c} 
      \toprule
       Number of planes                 &146 	\\  \midrule
       Number of poses                  &344    \\  \midrule
       Number of factors                &1974    \\ \midrule
	  \midrule
       CNN segmentation (ms)	     		&17.8		\\  \midrule
       Edge detection and selection (ms) &13.2		\\  \midrule
       Data association (ms)             &$<$1		\\  \midrule
       iSAM optimization (ms)            &17.4		\\  \midrule
       Total frame time (ms)             &\textbf{49.4}		\\  \bottomrule       
      \end{tabular}
 \end{center}
\end{table}

\subsection{Discussion}
\subsubsection{Height effect on map scale}
Unlike RGB-D plane SLAM whose plane measurements are generated by the actual depth sensor, our plane measurement and scale is determined by the camera height in the pop-up process in Section \ref{sec:pop_up_pose}. Camera height cannot be constrained by the plane measurements any more and therefore can only be constrained by other information such as odometry measurements or other sensors such as IMU. If other information is inaccurate or inaccessible, the map scale and camera height might drift using the plane SLAM alone. During all the experiments, we did not encounter the scale drift problem because the cameras are kept at a nearly constant height. In the future, we would like to integrate with other sensors.

%Unlike the RGB-D plane SLAM whose plane measurements are generated by the actual depth sensor, our plane measurement comes from the pop-up process whose scale is determined by the camera height in Section III-C. So the map scale cannot be constrained by the plane measurements anymore and there fore can only be constrained by other information such as odometry measurements or other sensors such as IMU

\subsubsection{Ground effect on graph complexity}
Since the pop-up process in Section \ref{sec:pop_up_pose} requires the ground plane to be visible, the ground plane landmark is connected to all the camera poses as shown in Figure \ref{fig:slam_graph}. This will reduce the sparsity of the information matrix and increase the computational complexity in theory. However, it can be alleviated by \textit{variable reordering} before matrix decomposition, for example using the COLAMD algorithm that will force this variable towards the last block column, thereby reducing its impact (see \cite{kaess2008isam}). From the experiments, the ground plane usually increases \textit{fill-in} (added non-zero entries) by only about 10\%.

%\section{Analysis}
%\label{sec:analysis}
%\input{analysis}

\section{Conclusions}
\label{sec:conclusions}
% propose what... results... future  
In this paper, we propose \textit{Pop-up Plane SLAM}, a real-time monocular plane SLAM system 
combined with scene layout understanding. It is especially suitable for low-texture environments 
because it can generate a rough 3D model even from a single image. 
We first extend previous work to pop up a 3D plane world from a single image by detecting 
the ground-wall edges and estimating camera rotation. Then, we formulate a plane 
SLAM approach to build a consistent plane map across multiple frames and also provide good state estimates. 
The plane landmark measurement comes from the single image pop-up model. 
We utilize the minimal plane representation for optimization and also implement plane SLAM loop closing.

%and continuously updated using the latest SLAM optimized camera poses.

Since plane SLAM itself is easily under-constrained in some environments, we propose 
to combine it with point based LSD SLAM in two ways: the first is \textit{Depth Enhanced LSD SLAM} 
by integrating pop-up pixel depth into LSD depth estimation, the second is \textit{LSD Pop-up SLAM}, which 
uses poses from \textit{Depth Enhanced LSD SLAM} as odometry constraints and runs a separate \textit{Pop-up Plane SLAM}.

In the experiment with the public TUM dataset, \textit{Pop-up Plane SLAM} generates a dense 3D map with depth 
error of $6.2$ cm and state estimates error of $3.9\%$ while the state-of-art LSD or ORB SLAM both fail.
Two collected large corridor datasets are also used to demonstrate its practicality and advantages. The loop closure 
error of \textit{LSD Pop-up SLAM} on a $60$m long dataset is only $0.67\%$, greatly outperforming LSD and 
ORB SLAM methods. The runtime analysis demonstrates that our algorithm could run in (near) real-time over $10$Hz.

In the future, we want to combine point, edge and plane landmarks in a unified SLAM framework. 
In addition, we would like to test this algorithm on robots. Besides, more work needs to be done 
in clutter corridors where ground-wall boundaries may be occluded.

\section*{Appendix}
\label{sec:appendix}
\subsection{Submodular Edge Selection}
Here, we briefly provide the optimality analysis of the edge selection as an extension to Section 
\ref{sec:Optimal boundary detection}. We prove that it is a submodular set selection problem
with matroid constraints \cite{krause2012submodular}.

\subsubsection{Monotonicity}
The score function $F$ in Equation \eqref{eq:edge cost function} is obviously monotonically increasing because
adding more edges, the covering in image horizontal direction will not decrease.

\subsubsection{Submodularity}
We first define the marginal gain of $e$ \textit{wrt.} $S$ as the increase of score $F$ after adding element $e$ into $S$, namely
$$\bigtriangleup(e\mid S) := F(S \cup{ \lbrace e \rbrace}) - F(S)$$

For two sets $S_1 \subset S_2$, edge $e$ may overlap with more edges in $S_2$ and thus reduce the marginal gain compared to
$S_1$, so it satisfies the submodularity condition:
$$ \bigtriangleup(e\mid S_1) \geq  \bigtriangleup(e\mid S_2), \ \forall S_1 \subseteq
S_2 $$
%It is a special case of set covering problem [submodular function maximization, tractability: practical
%approaches.] in 1D.

\subsubsection{Matroid constraint type}
We can remove the edges that are far from CNN boundary before submodular optimization, so we only
consider the second constraint $I_{ovlp}$ in Equation \eqref{eq:edge constraint 2}. Denote all the conflicting edge 
pairs as $ E_i=\lbrace (e_{i1},e_{i2}) \mid O(e_{i1},e_{i2}) \geq \delta_{ovlp} \rbrace, i=1,2,...,k$. 
 For each $E_i$, we form a partition of the ground set $V$ by two disjoint sets 
$P_i=\lbrace E_i,V \setminus E_i \rbrace$ and thus can form a partition matroid constraint $I_{i}^m=\lbrace S\colon | S \cap P_i^1 | \leq 1, | S \cap P_i^2 | \leq n \rbrace$, where $P_i^1$ and $P_i^2$ are two elements of $P_i$.
This is because we can pick at most one element from $E_i$. The union of $k$ such separate matroid constraints 
forms the original constraint $I_{ovlp}=I_{1}^m \cap I_{2}^m... \cap I_{k}^m$.

\subsubsection{Optimality}
From \cite{krause2012submodular}, the greedy algorithm in Equation \eqref{eq:greedy} of 
the submodular optimization with matroid constraints is guaranteed to produce a solution $S$ such that
$F(S) \geq \frac{1}{k+1} \max_{S \subseteq I} F(S)$. It is also important to note that this is only
a worst case bound and in most cases, the quality of solution obtained will be much better than this lower bound.

\addtolength{\textheight}{-12cm}   % This command serves to balance the column lengths
                                  % on the last page of the document manually. It shortens
                                  % the textheight of the last page by a suitable amount.
                                  % This command does not take effect until the next page
                                  % so it should come on the page before the last. Make
                                  % sure that you do not shorten the textheight too much.

%%%%%%%%%%%%%%%%%%%%%%%%%%%%%%%%%%%%%%%%%%%%%%%%%%%%%%%%%%%%%%%%%%%%%%%%%%%%%%%%

\section*{ACKNOWLEDGMENTS}

This work was supported by NSF award IIS-1328930 and IIS-1426703, and by ONR grant N00014-14-1-0373.

%%%%%%%%%%%%%%%%%%%%%%%%%%%%%%%%%%%%%%%%%%%%%%%%%%%%%%%%%%%%%%%%%%%%%%%%%%%%%%%%

\bibliographystyle{unsrt}    % reference order according to appearance
\bibliography{ref}

\end{document}